\documentclass[sigconf, screen]{acmart}
\AtBeginDocument{%
  }

\setcopyright{acmlicensed}
\copyrightyear{2026}
\acmYear{2026}
\acmDOI{XXXXXXX.XXXXXXX}

\renewcommand\footnotetextcopyrightpermission[1]{}

\acmISBN{978-1-4503-XXXX-X/2018/06}

\usepackage{multirow}
\usepackage{makecell}
\usepackage{pifont} 
\usepackage{xcolor} 
\usepackage[table]{xcolor}
\usepackage{subcaption}

\newcommand{\cmark}{\textcolor{green!70!black}{\ding{52}}} 
\newcommand{\xmark}{\textcolor{red}{\ding{56}}}          
\begin{document}

\title{Decoding the Delta: Unifying Remote Sensing Change Detection and Understanding with Multimodal Large Language Models}
\author{Xiaohe Li}
\email{lixiaohe@aircas.ac.cn}
\affiliation{
\city{Beijing}
\country{China}
\institution{Aerospace Information Research Institute, CAS}
}
\author{Jiahao Li}
\email{lijiahao243@mails.ucas.ac.cn}
\affiliation{
\city{Beijing}
\country{China}
\institution{Aerospace Information Research Institute, CAS}
}
\author{Kaixin Zhang}
\email{zhangkaixin25@mails.ucas.ac.cn}
\affiliation{
\city{Beijing}
\country{China}
\institution{Aerospace Information Research Institute, CAS}
}
\author{Yuqiang Fang}
\email{fangyuqiang@nudt.edu.cn}
\affiliation{
\city{Beijing}
\country{China}
\institution{Space Engineering University}
}
\author{Leilei Lin}
\email{leilei_lin@cnu.edu.cn}
\affiliation{
\city{Beijing}
\country{China}
\institution{Capital Normal University}
}
\author{Hong Wang}
\email{wanghong@aircas.ac.cn}
\affiliation{
\city{Beijing}
\country{China}
\institution{Aerospace Information Research Institute, CAS}
}
\author{Haohua Wu}
\email{wuhaohua23@mails.ucas.ac.cn}
\affiliation{
\city{Beijing}
\country{China}
\institution{Aerospace Information Research Institute, CAS}
}
\author{Zide Fan}
\email{fanzd@aircas.ac.cn}
\affiliation{
\city{Beijing}
\country{China}
\institution{Aerospace Information Research Institute, CAS}
}

\renewcommand{\shortauthors}{Trovato et al.}

\begin{abstract}
 While Multimodal Large Language Models (MLLMs) excel in general vision-language tasks, their application to remote sensing change understanding is hindered by a fundamental "temporal blindness". Existing architectures lack intrinsic mechanisms for multi-temporal contrastive reasoning and struggle with precise spatial grounding. To address this, we first introduce Delta-QA, a comprehensive benchmark comprising 180k visual question-answering samples. Delta-QA unifies pixel-level segmentation and visual question answering across bi- and tri-temporal scenarios, structuring change interpretation into four progressive cognitive dimensions. Methodologically, we propose Delta-LLaVA, a novel MLLM framework explicitly tailored for multi-temporal remote sensing interpretation. It overcomes the limitations of naive feature concatenation through three core innovations: a Change-Enhanced Attention module that systematically isolates and amplifies visual differences, a Change-SEG module utilizing Change Prior Embedding to extract differentiable difference features as input for the LLM, and Local Causal Attention to prevent cross-temporal contextual leakage. Extensive experiments demonstrate that Delta-LLaVA decisively outperforms leading generalist MLLMs and specialized segmentation models in complex change deduction and high-precision boundary localization, establishing a unified framework for earth observation intelligence.
\end{abstract}

\begin{teaserfigure}
   \includegraphics[width=\textwidth]{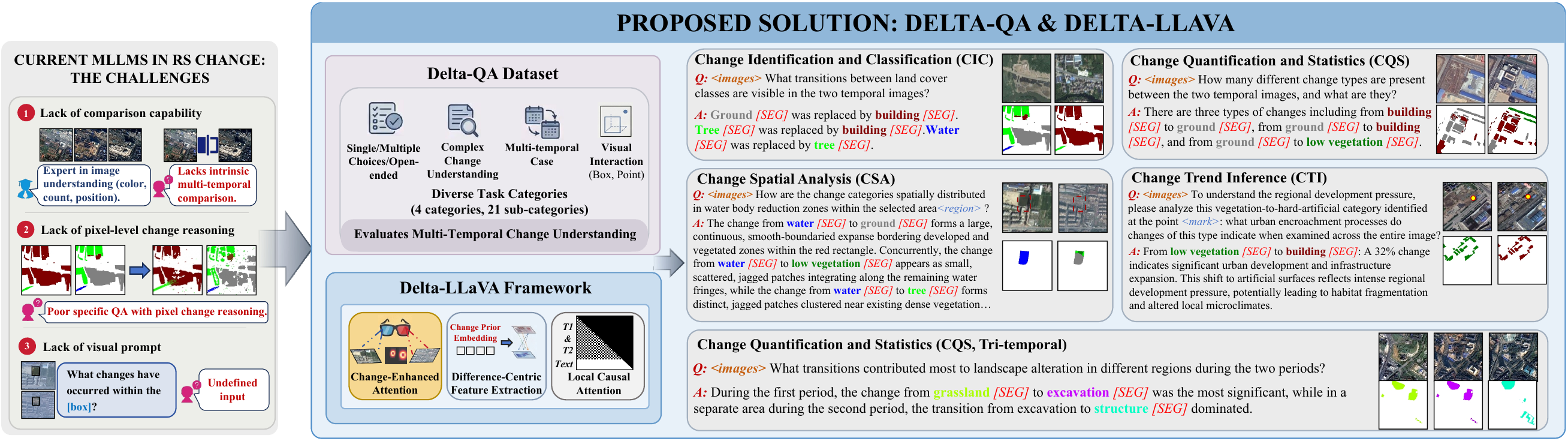}
   \caption{Our work primarily addresses the unified task of multi-temporal remote sensing change detection and understanding. The proposed Delta-QA dataset encompasses various QA task types and interaction modalities.}
   \label{fig:first_page}
\end{teaserfigure}
\maketitle
\begin{table*}[tpb]
\centering
\caption{Comparison of existing multi-temporal remote sensing datasets for change perception and understanding.}
\vspace{-0.2cm}
\renewcommand{\arraystretch}{0.9}
\label{tab:dataset_comparison}
\large
\resizebox{0.95\textwidth}{!}{
\begin{tabular}{l|l|cccccccc}
\toprule
\textbf{Category} & \textbf{Dataset} & \textbf{\makecell{Frame \\ Number}} & \textbf{\makecell{Land-Cover  Type}}& \textbf{\makecell{Containing\\Pixel Mask}} & \textbf{\makecell{Textual \\ QA/Caption}} & \textbf{Task Type} & \textbf{\makecell{Interactive \\ Prompt}} & \textbf{Resolution} & \textbf{\makecell{Annotation \\ Scale}} \\
\midrule
\multirow{5}{*}{\makecell{Semantic Change\\Detection}} 
& SECOND~\cite{yang2021asymmetric}        & 2 & 6 & \cmark & \xmark & Mask & \xmark & 0.5-3 & 4662 Image Pairs\\
& Landsat~\cite{yuan2022transformer}       & 2 & 4& \cmark & \xmark & Mask & \xmark & 30    & 8468 Image Pairs\\
& WUSU~\cite{yuan2022transformer}          & 3 & 11 & \cmark & \xmark & Mask & \xmark & 1     & 1116 Image Pairs\\
& LevirSCD~\cite{zhang2025foba}            & 2 & 16 & \cmark & \xmark & Mask & \xmark & 1-2   & 3225 Image Pairs\\
& JL1~\cite{jl1_scd_dataset}                                     & 2 & 5 & \cmark & \xmark & Mask & \xmark & 0.75  & 6000 Image Pairs\\
\midrule
\multirow{8}{*}{\makecell{Change Detection\\and Understanding}}
& SECOND-CC~\cite{karaca2025robust}        & 2 & 6 & \cmark & \cmark & Caption & \xmark & 0.5-3 & 30205 Captions \\
& RSCC~\cite{chen2025rscc}                 & 2 & - & \xmark & \cmark & Caption & \xmark & -     & 62351 Captions \\
& LEVIR-CC~\cite{liu2022remote}                                & 2 & - & \xmark & \cmark & Caption & \xmark & 0.5   & 50385 Captions \\
& CC-Foundation~\cite{wang2024ccexpert}& 2 & - & \xmark & \cmark & Caption& \xmark & - & 135k Captions \\
& LEVIR-MCI~\cite{liu2024change}                               & 2 & 2 & \cmark & \cmark & QA or Mask& \xmark & 0.5 & 50.3k Instructs\\
& ChangeChat~\cite{deng2025changechat}                               & 2 & 2 & \xmark& \cmark & QA& \xmark & 0.5 & 87k Instructs \\
& DVL-Bench~\cite{xuan2025dynamicvl}   & 5-10 & 5 & \cmark & \cmark & Unified QA and Simple Mask Positioning & Box only & 1 & 69926 Instructs \\
& Delta-QA (ours)                     & 2-3 & 15 & \cmark & \cmark & Unified QA and Complex Mask Reasoning& Point \& Box & 0.5-30 & 180k Instructs \\
\bottomrule
\end{tabular}
}
\vspace{-0.1cm}
\end{table*}
\section{Introduction}
Driven by rapid advancements in Earth observation, modern remote sensing systems continuously generate dense multi-temporal imagery, providing critical records for deciphering global dynamics such as urbanization and disaster response \cite{rahnemoonfar2023rescuenet,zhang2024earthgpt}. Concurrently, Multimodal Large Language Models (MLLMs) \cite{hu2025rsgpt,yin2024survey} have demonstrated extraordinary proficiency in vision-language alignment and zero-shot reasoning. Integrating MLLMs into remote sensing holds the potential to transcend the semantic bottlenecks of traditional task-specific computer vision, promising a leap toward open-ended and highly generalizable Earth observation intelligence \cite{Yao2025RemoteSAMTS}.

Despite these promising prospects, contemporary MLLMs remain fundamentally ill-equipped for the nuanced demands of remote sensing change understanding due to three critical limitations. First, existing works predominantly fine-tune current vision-language models to better understand intrinsic image content, specifically targeting improvements in color, count, and spatial positioning capabilities~\cite{luo2025large,wang2026geoeyes,li2026co, yang2026geoalignclip,kashyap2026bi}. They lack intrinsic mechanisms for multi-temporal contrastive reasoning, rendering them effectively "temporally blind" to evolutionary dynamics \cite{irvin2024teochat,noman2025cdchat}. Therefore, a profound interpretation and detailed analysis of intricate change dynamics are difficult to attain~\cite{wang2024ringmogpt,deng2026deltavlm}. Second, they operate at a coarse semantic level, lacking the spatial grounding requisite for precise change delineation~\cite{zhang2025georsmllm}. Custom integrations that attempt to graft spatial decoders or naively concatenate multitemporal features are fraught with tradeoffs, often degrading generalized reasoning or failing to achieve rigorous spatiotemporal alignment at the pixel scale \cite{ma2025geomag}. Finally, practical interpretation demands flexible and multidimensional interaction modes, yet current structurally monolithic approaches support only rigid interactions~\cite{zhang2025unichange,huang2025reasoncd}. Consequently, there is an urgent need for a unified framework that can seamlessly integrate text, spatial prompts, and precise mask outputs to empower diverse analytical workflows.

To overcome these foundational limitations, this paper introduces synergistic advancements in both the data foundation and the model architecture. By integrating natural language with segmentation masks, we seek to empower vision-language models to genuinely comprehend the temporal variations in remote sensing imagery, effectively capturing the 'Delta' between them.

On the data front, we propose an automated pipeline for generating tasks for change perception and reasoning and introduce the \textbf{Delta-QA} dataset. This comprehensive benchmark synthesizes existing datasets, namely SECOND~\cite{yang2021asymmetric}, Landsat-SCD~\cite{yuan2022transformer}, and WUSU~\cite{shi2023multi}, to support bitemporal and tritemporal change scenarios at the pixel scale. It encompasses four principal task categories: Change Identification and Classification (\texttt{[CIC]}), Change Quantification and Statistics (\texttt{[CQS]}), Change Trend Inference (\texttt{[CTI]}), and Change Spatial Analysis (\texttt{[CSA]}). Furthermore, the dataset spans multiple-choice and open-ended formats and natively accommodates versatile input modalities, including spatial points, bounding boxes, and holistic image queries.

Methodologically, we propose \textbf{Delta-LLaVA}, a novel MLLM framework tailored for multi-temporal remote sensing interpretation. Building upon token-based alignment techniques \cite{zhang2024omg, lai2024lisa}, Delta-LLaVA unifies precise change detection with textual and visual prompt reasoning. To overcome the severe information loss inherent in naive feature concatenation, we introduce three core architectural innovations: (i) \textbf{Change-Enhanced Attention (CEA)} bidirectionally amplifies salient structural variations and systematically suppresses background noise to distill pure change signals. (ii) \textbf{Change-SEG}, powered by a Change Prior Embedding, bridges the visual latent spaces of paired images to extract shared difference representations, granting the model an intrinsic capacity for direct image comparison. (iii) \textbf{Local Causal Attention (LCA)} confines visual token interactions strictly within their respective temporal phases, preventing cross-temporal feature leakage and ensuring unbiased processing. Optimized via a two-stage strategy of visual pretraining and supervised fine-tuning, Delta-LLaVA seamlessly integrates natural language generation with mask decoding to deliver both nuanced textual analyses and highly accurate segmentation masks. Our source code and dataset will be open-sourced to the community.

Our contributions are threefold:
\begin{itemize}
    \item We introduce \textbf{Delta-QA}, a dataset of 180k multi-temporal QA samples that structures fine-grained change understanding into four cognitive dimensions (CIC, CQS, CTI, CSA), effectively bridging pixel-level segmentation and visual QA.
    \item We propose \textbf{Delta-LLaVA}, a novel MLLM tailored for multi-temporal change understanding. It explicitly extracts and amplifies temporal difference priors while preventing cross-temporal feature confounding.
    \item Delta-LLaVA achieves state-of-the-art performance on Delta-QA and DVL-Bench~\cite{xuan2025dynamicvl}, decisively outperforming leading baselines in both complex change reasoning and precise spatiotemporal mask delineation.
\end{itemize}
\section{Related Work}
\subsection{MLLMs in Remote Sensing}
Early Remote Sensing MLLMs (e.g., GeoChat~\cite{kuckreja2024geochat}, RSUniVLM~\cite{liu2024rsunivlm}, GEOBench-VLM~\cite{danish2025geobench}) primarily focus on single-image understanding, restricting spatial grounding to coarse bounding boxes. To capture temporal dynamics, recent research has introduced sequence models such as EarthDial~\cite{soni2025earthdial} and bi-temporal change models such as BTCChat~\cite{li2025btcchat}, alongside advanced change captioning benchmarks (e.g., SECOND-CC~\cite{karaca2025robust}, RSCC~\cite{chen2025rscc}, LEVIR-CC~\cite{liu2022remote}, CC-Foundation~\cite{wang2024ccexpert}). However, these models predominantly output text or boxes, fundamentally lacking pixel-level mask alignment.

To address this coarse-grained limitation, the "reasoning segmentation" paradigm (e.g., LISA~\cite{lai2024lisa}, OMG-LLaVA~\cite{zhang2024omg}) has been adapted to generate pixel-level masks, with models such as LISAT~\cite{quenum2025lisat} applying this directly to remote sensing imagery. Nevertheless, these architectures are inherently optimized for static single images. Without dedicated spatiotemporal mechanisms to explicitly align and compare cross-temporal features, they remain inadequate for complex dynamic interpretation and true change understanding.

\begin{figure*}[t]
  \centering
  \includegraphics[width=0.89\textwidth]{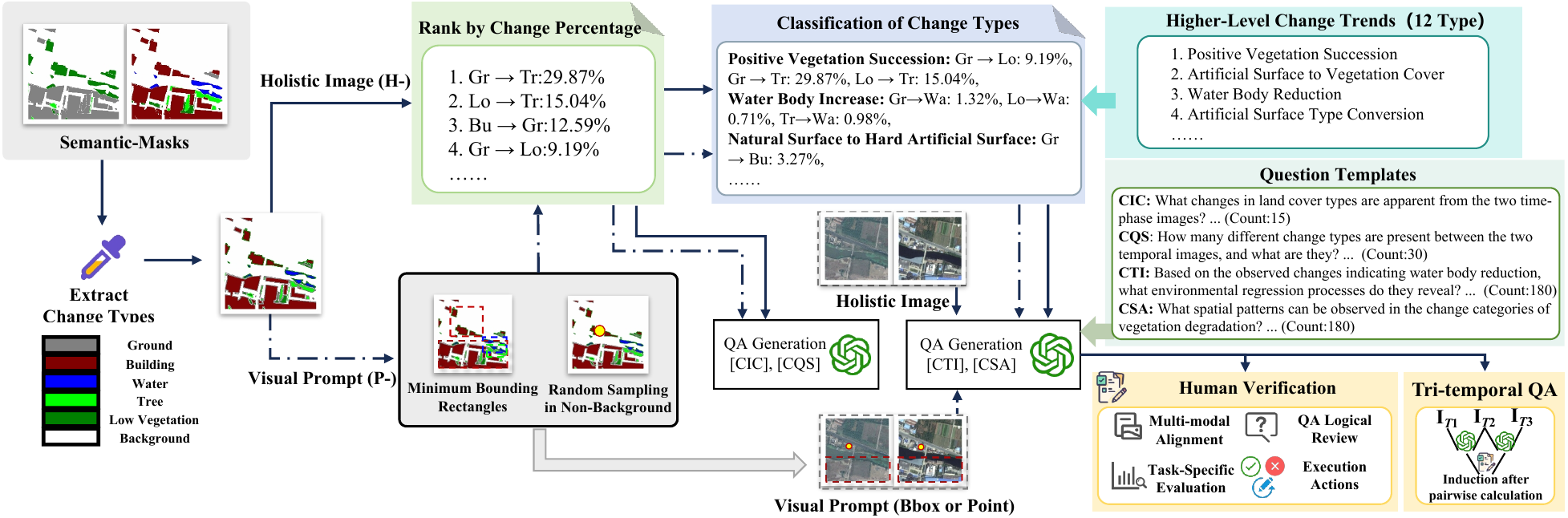}
  \vspace{-0.1cm}
  \caption{An overview of the data annotation framework for the Delta-QA dataset.}
  \label{fig:dataset_pipeline}
  \vspace{-0.2cm}
\end{figure*}

\subsection{Semantic Change Detection in Remote Sensing}
Semantic change detection (SCD) aims to identify category transitions of ground objects. Recent vision-only methods have significantly advanced spatiotemporal modeling, boundary delineation, and generalization~\cite{chen2024changemamba, ding2024joint, zhang2025foba, mei2024scd}. Despite achieving high-precision pixel-level alignment on mainstream benchmarks~\cite{yang2021asymmetric,yuan2022transformer,shi2023multi,zhang2025foba,jl1_scd_dataset}, they are restricted to a closed-set paradigm, fundamentally lacking the ability to process open-ended instructions or perform multimodal reasoning. Recent interactive works in remote sensing, like ChangeChat~\cite{deng2025changechat} and DeltaVLM~\cite{deng2026deltavlm}, are limited to pure text QA, where spatial localization relies solely on text coordinates. While ReasonCD~\cite{huang2025reasoncd} performs temporally implicit change detection, it necessitates explicit spatial grounding. UniChange~\cite{zhang2025unichange} integrates binary change detection (BCD) and SCD, but remains a purely discriminative model that lacks the capacity for change interpretation.

As summarized in Table~\ref{tab:dataset_comparison}, existing unified change datasets exhibit notable limitations. For instance, LEVIR-MCI~\cite{liu2024change} outputs text and masks independently, lacking dense alignment and support for point/box visual prompts. Although DVL-Bench~\cite{xuan2025dynamicvl} achieves synchronous outputs via coarse bounding boxes, its text merely provides category labels rather than actual change descriptions. To address these gaps, we introduce Delta-QA, which utilizes fine-grained point-and-box prompts to synchronously generate precise masks alongside detailed semantic descriptions.

\section{Delta-QA Dataset}
We introduce \textbf{Delta-QA}, a unified remote sensing dataset mapping multi-temporal visual inputs to multidimensional QA and pixel-level masks for fine-grained change evaluation. Derived from SECOND~\cite{yang2021asymmetric}, Landsat-SCD~\cite{yuan2022transformer}, and WUSU~\cite{shi2023multi}, it generates 180,876 QA samples supporting point and box visual prompts under original train/test splits. It comprises three subsets:
\begin{itemize}
    \item \textbf{Delta-SECOND}: 4,654 bi-temporal pairs (512×512, 0.3–5m). Yields 78,950 train and 43,640 test samples across six change classes plus one unchanged.
    \item \textbf{Delta-Landsat}: 2,385 bi-temporal pairs (416×416, 30m). Contains 44,088 train and 10,973 test samples covering four change categories and one unchanged.
    \item \textbf{Delta-WUSU}: 646 tri-temporal sequences (512×512, 1m). Provides 1,712 train and 1,513 test samples focusing on spatio-temporal urban dynamics (seven change categories, one unchanged).
\end{itemize}
We systematically structure the QA tasks into four progressive cognitive dimensions (Fig.~\ref{fig:first_page}): Change Identification and Classification (\texttt{[CIC]}), Change Quantification and Statistics (\texttt{[CQS]}), Change Trend Inference (\texttt{[CTI]}), and Change Spatial Analysis (\texttt{[CSA]}).

\subsection{Dataset Curation Pipeline}
Fig.~\ref{fig:dataset_pipeline} illustrates our automated data annotation pipeline. To mitigate visual hallucinations during fine-grained perception, we first extract land cover change types from temporal semantic segmentation masks, serving as the sole objective ground truth. The pipeline operates under two spatial scopes: a Holistic (H-) perspective targeting the entire image, and a Visual Prompt (P-) perspective utilizing dynamically extracted Minimum Bounding Rectangles (MBRs) or random points in non-background to restrict the interaction scope.

By extracting statistical metrics (e.g., instance count, area proportion) and abstracting foundational transitions into higher-level evolution trends, we integrate these features with spatial coordinates to prompt GPT-4o to generate four categories of Question-Answering (QA) tasks:
(i) \textbf{Change Identification and Classification (CIC)}: Focusing on comprehensive and specific source/target change identification.
(ii) \textbf{Change Quantification and Statistics (CQS)}: Requiring exact numerical calculations for change composition and area proportions.
(iii) \textbf{Change Trend Inference (CTI)}: Generating open-ended descriptions regarding the direction of land surface evolution.
(iv) \textbf{Change Spatial Analysis (CSA)}: Analyzing spatial distribution patterns of change events. 

Coupling these tasks with two spatial modes yields eight distinct QA sets. Crucially, all generated texts maintain a strict hard-mapping with corresponding pixel-level masks and undergo multi-stage human verification to ensure precise multi-modal alignment. Furthermore, we extend this pipeline to tri-temporal sequences by decoupling them into adjacent bi-temporal stages ($T_1 \rightarrow T_2$, $T_2 \rightarrow T_3$). GPT-4o concurrently processes both stages to construct ``composite evolution events,'' accurately capturing interleaved land-cover dynamics. Detailed extraction rules, verification criteria, and templates are in Appendix~A.

\begin{figure}[htbp]
  \centering
  \begin{minipage}{0.21\textwidth}
    \centering
    \includegraphics[width=\linewidth]{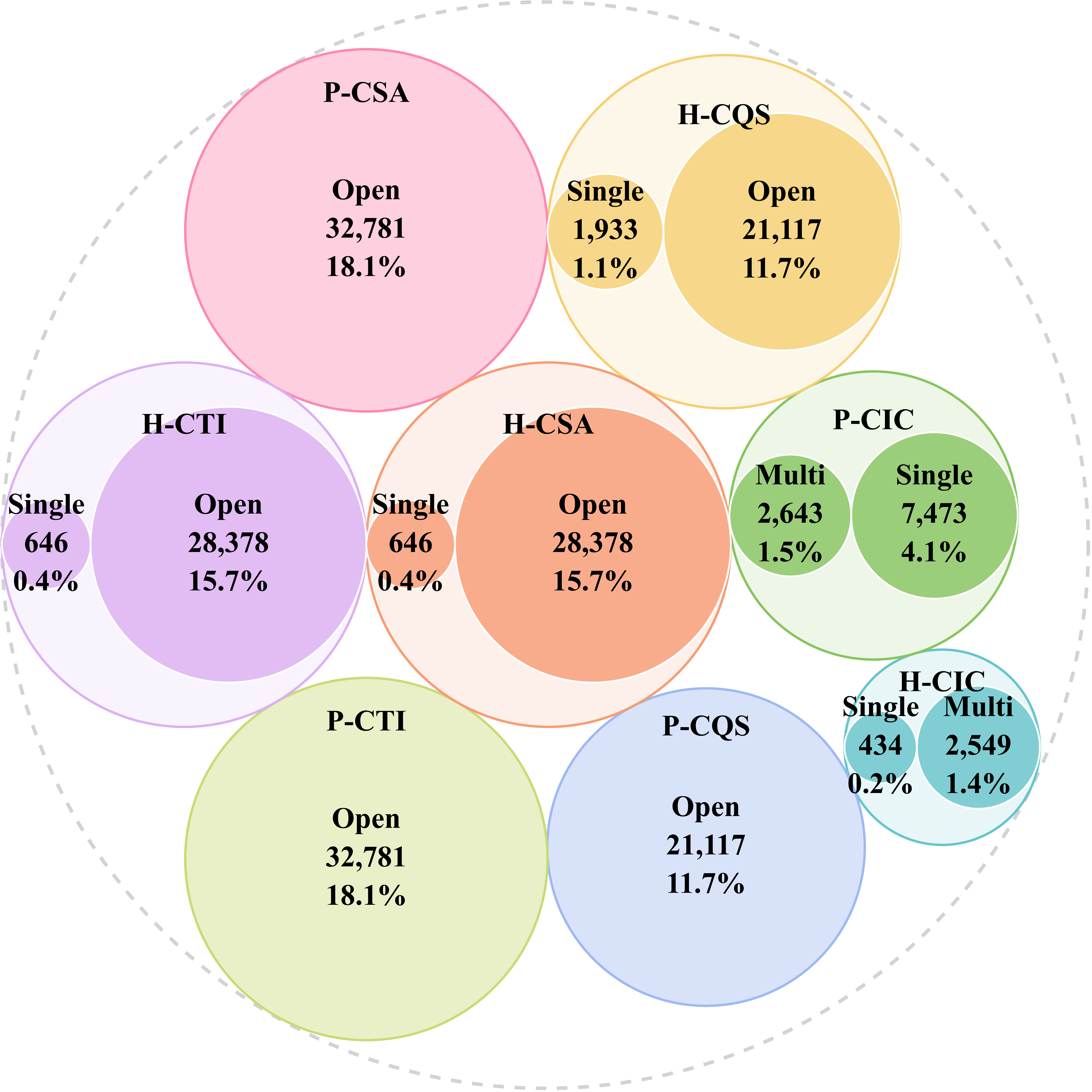}
    \caption{Types and question counts in Delta-QA.}
    \label{fig:all_data}
  \end{minipage}
  \hfill
  \begin{minipage}{0.21\textwidth}
    \centering
    \includegraphics[width=\linewidth]{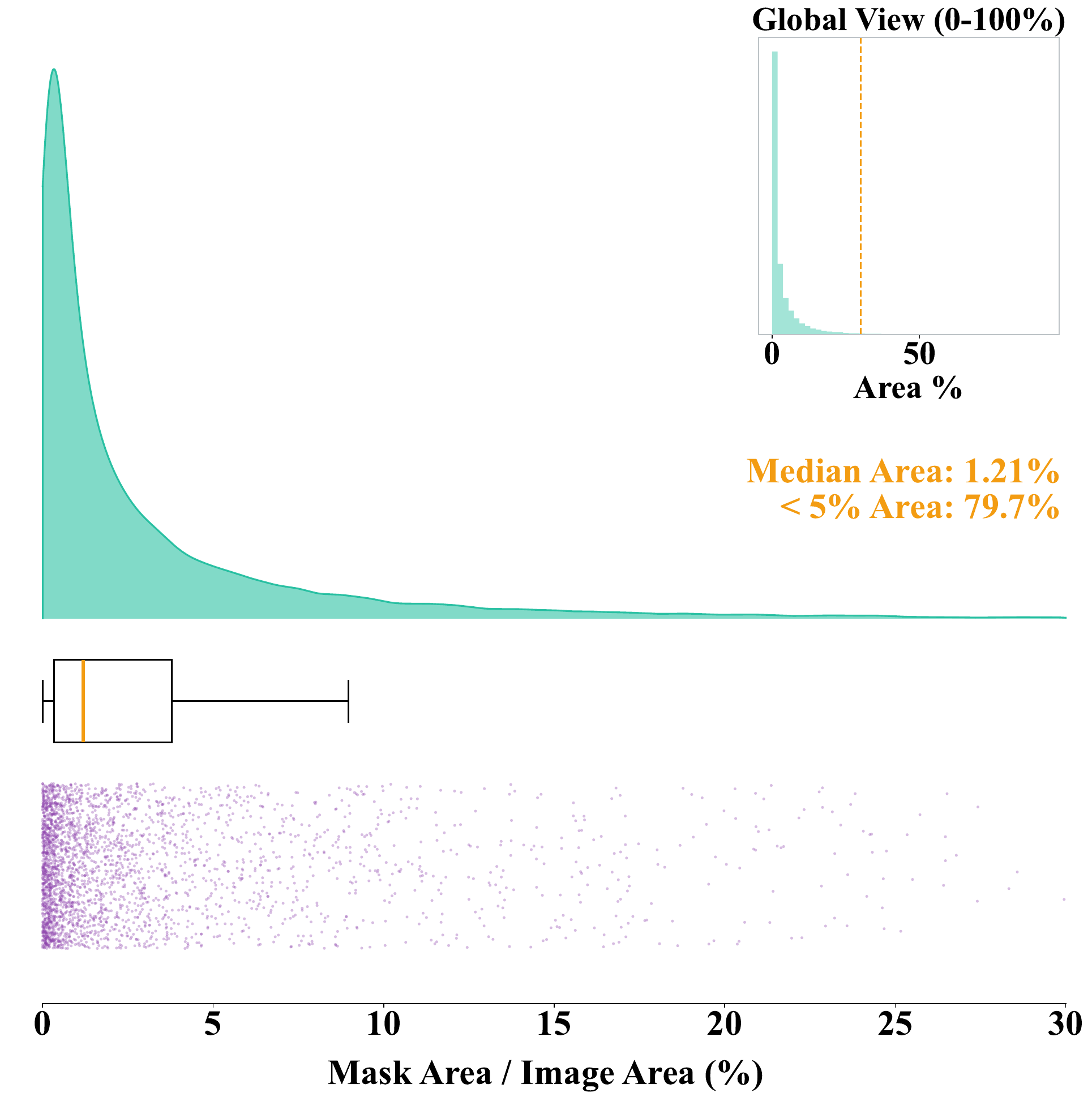}
    \caption{Mask-to-image ratio in Delta-SECOND subset.}
  \label{fig:combined_dataset_qa}
  \end{minipage}
      \vspace{-0.5cm}
\end{figure}

\begin{figure}[htbp]
  \centering
  \begin{minipage}{0.21\textwidth}
    \centering
    \includegraphics[width=\linewidth]{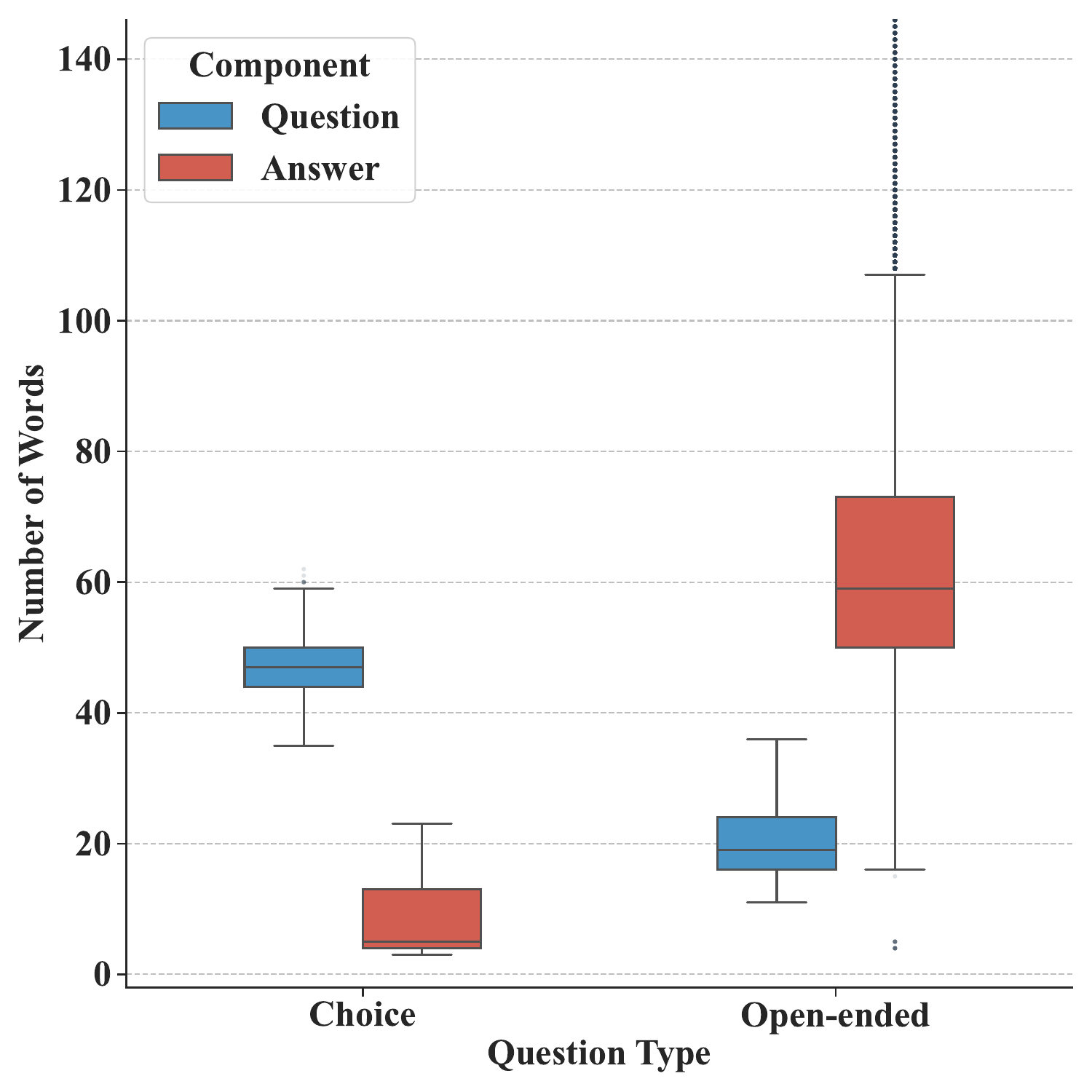}
    \caption{QA text length distribution in Delta-SECOND.}
    \label{fig:text_length}
  \end{minipage}
    \hfill
  \begin{minipage}{0.21\textwidth}
    \centering
    \includegraphics[width=\linewidth]{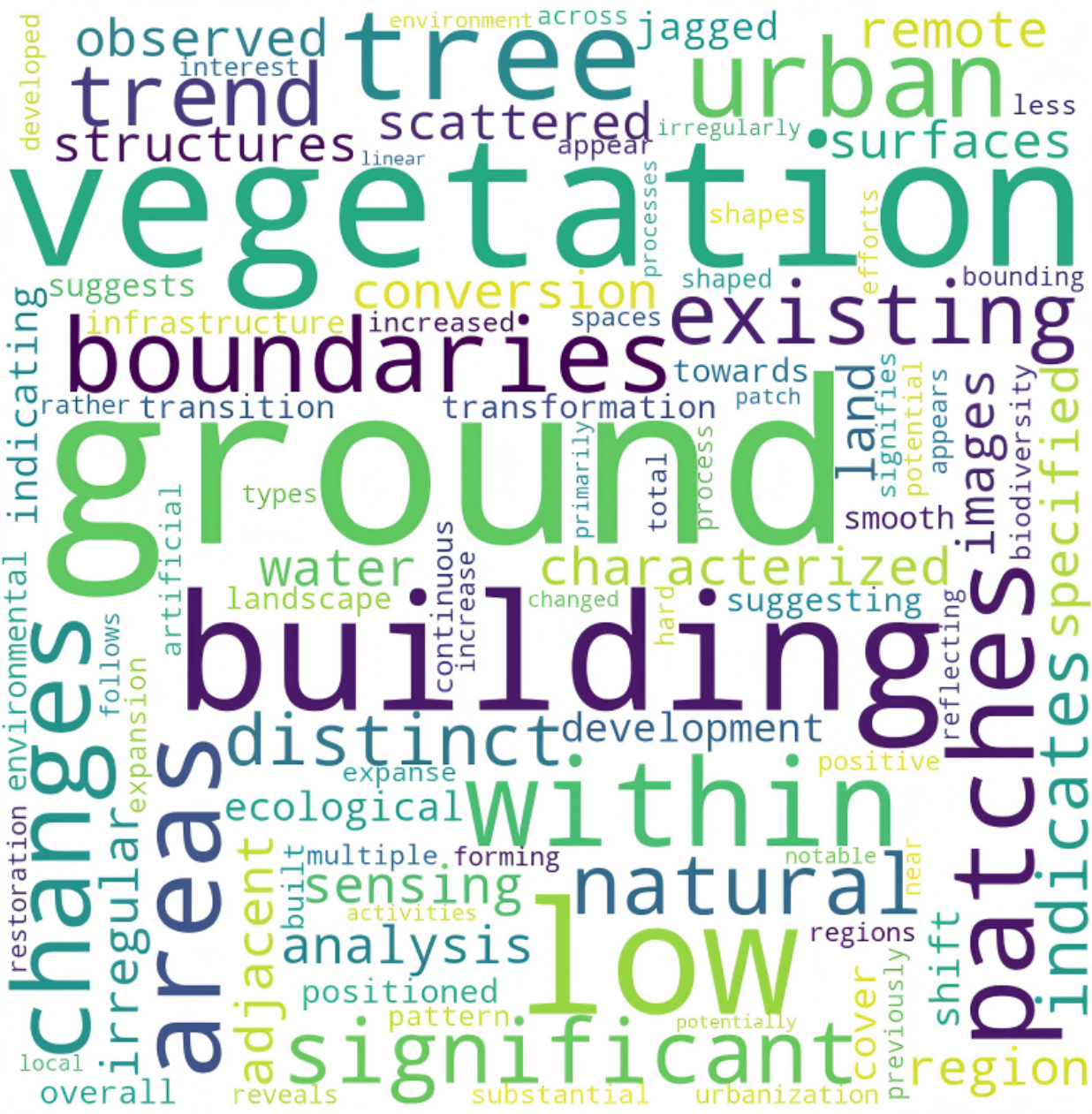}
    \caption{Word frequency cloud of Delta-SECOND.}
    \label{fig:word_cloud}
  \end{minipage}
      \vspace{-0.4cm}
\end{figure}
\begin{figure}[t]
  \centering
  \includegraphics[width=0.85\linewidth]{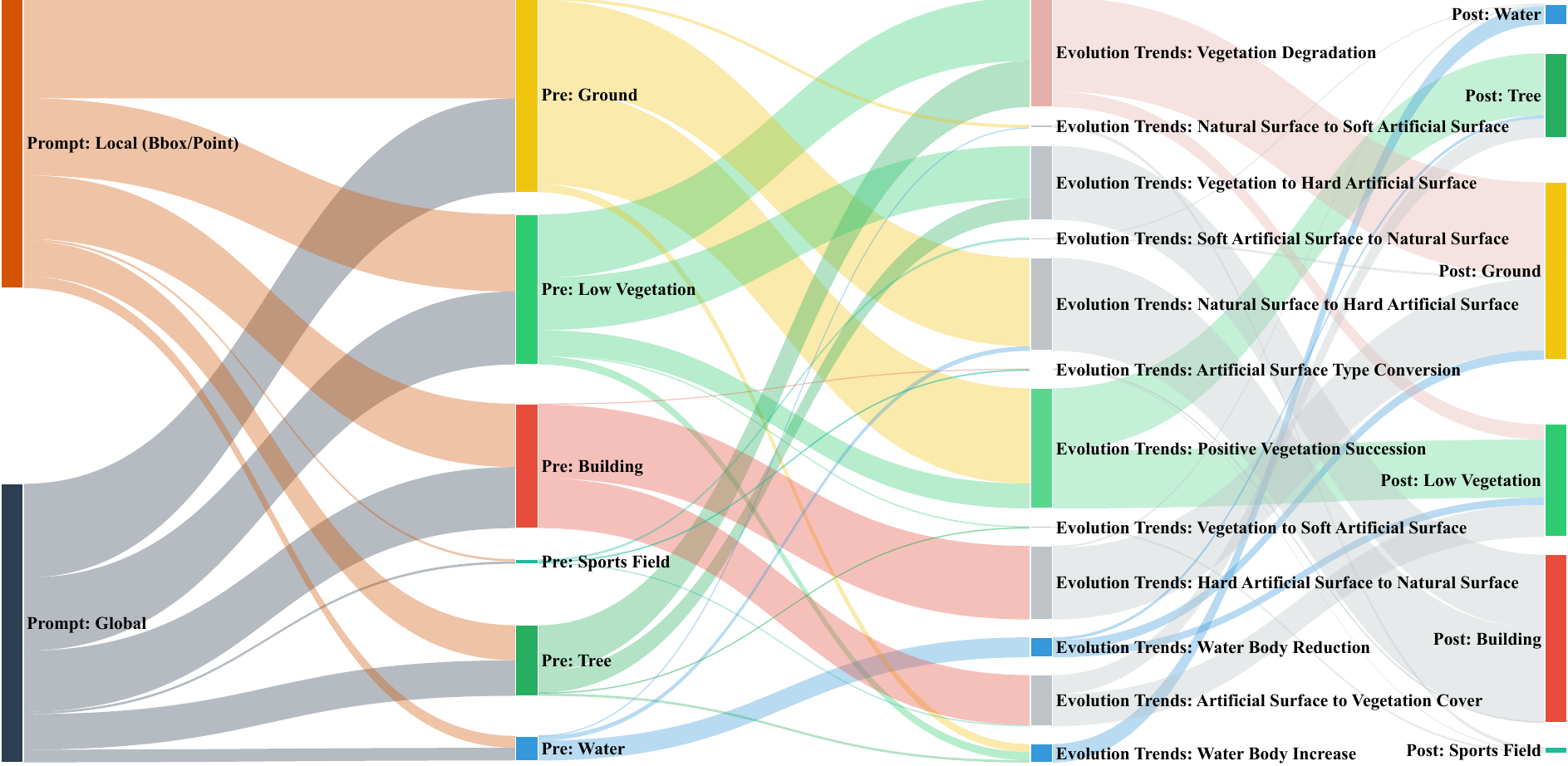}
  \caption{Land cover transitions and evolution trends in Delta-SECOND.}
  \label{fig:second_trends}
      \vspace{-0.5cm}
\end{figure}

\subsection{Dataset Statistics}
We quantitatively analyze Delta-QA's overall distribution and use the Delta-SECOND subset to detail spatial scales, textual characteristics, and land cover evolution (further subset details in Appendix~A). As shown in Fig.~\ref{fig:all_data}, Delta-QA comprises 180,876 multimodal QA pairs, split into 84,081 holistic and 96,795 visual prompt-guided samples. Task-wise, high-order open-ended reasoning (CQS, CTI, CSA) dominates with 167,777 samples (92.8\%), while Change Identification and Classification (CIC) provides 13,099 multiple-choice samples, emphasizing complex text generation. 

In the Delta-SECOND subset, visual instruction mask areas exhibit a right-skewed long-tail distribution (Fig.~\ref{fig:combined_dataset_qa}). With a median coverage of 1.21\% and 79.7\% of masks under 5\%, this spatial sparsity enforces strict visual grounding and prevents reliance on linguistic shortcuts. Textually (Fig.~\ref{fig:text_length}, \ref{fig:word_cloud}), multiple-choice answers have a median of 5 words, whereas open-ended responses average 60 words (tailing beyond 140). High-frequency words include foundational nouns (e.g., ``ground'', 172k) and analytical terms (e.g., ``trend'', 41k), confirming rich descriptions of spatial trends. Finally, Fig.~\ref{fig:second_trends} visualizes 53,554 land cover transitions. Macroscopic trends highlight vegetation succession (40.33\%), conversion to artificial surfaces (29.33\%), and ecological restoration (21.90\%). Micro-level paths are dominated by ``Ground $\rightarrow$ Building'' (15.55\%) and ``Building $\rightarrow$ Ground'' (12.58\%), authentically reflecting real-world geographic processes such as urban expansion.

\section{Delta-LLaVA Framework}
\begin{figure*}
  \includegraphics[width=\textwidth]{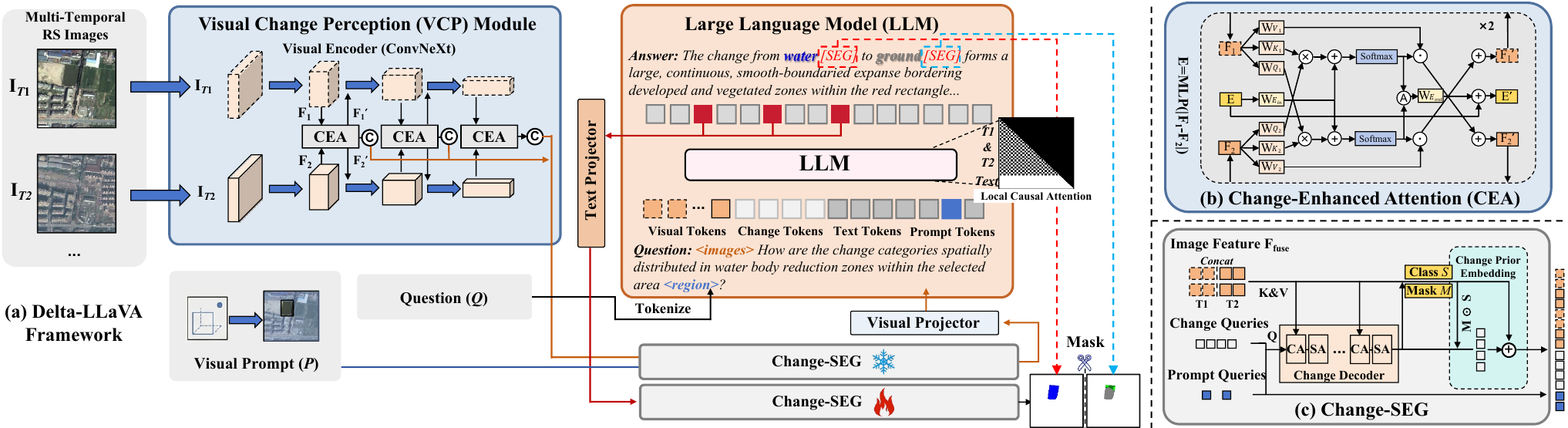}
\caption{Overall architecture of Delta-LLaVA. The VCP module explicitly amplifies differences in imagery feature. Visual, Change, and Prompt tokens generated by Change-SEG are concatenated with text tokens and fed into the LLM to jointly output textual answers and segmentation masks.}
    \vspace{-0.1cm}
  \label{fig:framework}
      \vspace{-0.3cm}
\end{figure*}
This paper proposes a unified vision-language framework for multi-temporal change detection and question-answering. Formally, the overall task is defined as:
\begin{equation}
    A_{t}, A_{m} = \text{LLM}(\{I_i\}_{i=1}^{K}, Q, P),
\end{equation}
where $\{I_i\}_{i=1}^{K}\in \mathbb{R}^{W\times H}$ denotes the input sequence of remote sensing images across $K$ temporal phases, $Q$ represents the change-related text query, and $P$ indicates an optional visual prompt. The model simultaneously outputs a corresponding textual answer $A_{t}$ and change detection mask indicated tokens $A_{m}$.

As illustrated in Fig.~\ref{fig:framework}, Delta-LLaVA consists of two primary components: a Visual Change Perception (VCP) module and a Large Language Model (LLM) optimized for change semantic understanding. The VCP module, incorporating a Change-Enhanced Attention (CEA) mechanism and a Change-SEG decoder, encodes the multi-temporal images and visual prompts into visual and change-indicator tokens. The LLM processes these tokens alongside text instructions to jointly generate a textual response and specialized change segmentation tokens (e.g., \texttt{[SEG]}). Finally, the decoder resolves these segmentation tokens into a precise, pixel-level change detection mask.

\subsection{Change-Enhanced Attention}
To extract multi-temporal features, the VCP module employs a ConvNeXt-L CLIP encoder~\cite{liu2022convnet} alongside a change-enhanced attention mechanism. Adjacent temporal images $I_{T_i}$ and $I_{T_{i+1}}$ (resized to $1024 \times 1024$) are processed independently and, after feature extraction and downsampling, mapped into 256 ($16 \times 16$) visual tokens per image, yielding $l$-th layer features $\{F^{(l)}_{I_{T_i}}, F^{(l)}_{I_{T_{i+1}}}\} \in \mathbb{R}^{W \times H \times d_f}$. To mitigate external perturbations common in remote sensing (e.g., illumination variations), we compute the first-order absolute distance for stages 2 to 4 as direct inputs to the segmentation decoder:
\begin{equation}
    F^{(l)}_{\text{diff}} = |F^{(l)}_{I_{T_i}} - F^{(l)}_{I_{T_{i+1}}}|.
\end{equation}

Subsequently, a Multi-Layer Perceptron (MLP) extracts the learned change-enhanced features $E_{\text{diff}} = \text{MLP}(F^{(l)}_{\text{diff}})$. Inspired by Edge-Augmented Graph Transformers~\cite{hussain2022global}, $E_{\text{diff}}$ acts as an affinity measure between collocated pixels, serving to enhance the representation differences between adjacent contents. Conceptually analogous to the stereoscopic polarizing glasses used in 3D cinema, which selectively filter distinct optical inputs for each eye to facilitate depth perception in the brain, this mechanism systematically isolates and amplifies the salient physical differences between paired temporal views.

Specifically, treating the visual features of the two adjacent images as symmetric source and target representations, we compute a point-wise cross-attention to capture strict spatial correspondences:
\begin{align}
    A_1^{(l)} &= \text{Softmax}\left(\frac{\langle F^{(l)}_{I_{T_i}}W_{Q_1}, F^{(l)}_{I_{T_{i+1}}}W_{K_2} \rangle}{\sqrt{d_f}} + E_{\text{diff}}W_{E_{in}}\right), \\
    A_2^{(l)} &= \text{Softmax}\left(\frac{\langle F^{(l)}_{I_{T_i}}W_{Q_2}, F^{(l)}_{I_{T_{i+1}}}W_{K_1} \rangle}{\sqrt{d_f}} + E_{\text{diff}}W_{E_{in}}\right),
\end{align}
where $\langle \cdot, \cdot \rangle$ denotes the pixel-wise inner product computed along the channel dimension. The matrices $\{W_{Q_1}, W_{Q_2}, W_{K_1}, W_{K_2}\} \in \mathbb{R}^{d_f \times d_f}$ and $W_{E_{in}} \in \mathbb{R}^{d_f \times 1}$ are learnable weight matrices used for linear projections. The features of the two temporal phases are then updated using the spatial attention maps $\{A_1^{(l)}, A_2^{(l)}\} \in \mathbb{R}^{W \times H}$ as follows:
\begin{equation}
    F^{(l)'}_{I_{T_i}} = F^{(l)}_{I_{T_i}} + A_1^{(l)} \odot (F^{(l)}_{I_{T_{i+1}}}W_{V_2}), \quad 
    F^{(l)'}_{I_{T_{i+1}}} = F^{(l)}_{I_{T_{i+1}}} + A_2^{(l)} \odot (F^{(l)}_{I_{T_i}}W_{V_1}),
\end{equation}
where $\odot$ denotes element-wise multiplication broadcasted along the channel dimension. 

This mechanism leverages cross-attention across multiple feature scales to effectively enhance pixel-level image differences, highlighting salient cross-temporal changes. Furthermore, we implement a two-layer network design where the edge features in the second layer are accumulated using information from the first layer, formulated as:
\begin{equation}
    E_{\text{diff}} = E_{\text{diff}} + \frac{(A_1^{(l)} + A_2^{(l)})}{2}W_{E_{out}},
\end{equation}
where $W_{E_{out}} \in \mathbb{R}^{1 \times d_f}$ projects the aggregated spatial attention back to the feature dimension to ensure superior change enhancement.

\subsection{Difference-Centric Token Extraction}

Existing large multimodal models primarily focus on understanding individual images or video content, lacking tailored designs for fine-grained image comparison. To establish intrinsic image comparison capabilities, we design a visual token extraction mechanism that is sensitive to multi-temporal differences. As illustrated in Fig.~\ref{fig:framework}, we propose the Change-SEG module. Inspired by OMG-Seg~\cite{zhang2024omg}, it introduces learnable difference-centric queries $Q_d \in \mathbb{R}^{N_d \times C}$ to automatically capture and highlight object-level change features. Taking the bi-temporal case as an example, to facilitate processing by the Mask2Former~\cite{cheng2022masked} architecture and enable simultaneous observation of all temporal phases for comprehensive change understanding, multi-scale VCP features (stages 2 to 4) are spatially concatenated into a unified representation $F_{\text{fuse}} \in \mathbb{R}^{W \times 2H \times C}$. This serves as Key and Value in a multi-layer attention decoder to update $Q_d$. Additionally, encoded visual prompts (e.g., points or boxes) generate targeted queries and attention masks~\cite{kirillov2023segment} to direct regional focus.

As illustrated in Fig.~\ref{fig:framework}, to further inject the global difference prior into the multi-temporal features, we introduce the Change Prior Embedding (CPE) strategy. $Q_d$ interacts with stage 1 features to predict a foreground semantic mask $M \in \mathbb{R}^{W \times 2H \times N_d}$ and category scores $S \in \mathbb{R}^{1 \times N_d}$. This dynamically modulates the spatial fusion ratio to enhance visual features:
\begin{equation}
    T_v = \text{Softmax}(M \odot S) Q_d + F_{\text{fuse}},
\end{equation}
where $\odot$ denotes element-wise multiplication with channel-wise broadcasting, and Softmax is applied along $N_d$. Unlike OMG-Seg, our $Q_d$ shares a receptive field across temporal features, acting as a seamlessly integrated differentiated prior. Queries $Q_d$ with high weights effectively activate in response to the changed regions. 

To balance general object semantics with specific change adaptation, Change-SEG employs a dual-branch strategy with identical pre-trained initialization. The primary branch remains frozen to prevent catastrophic forgetting, while the secondary branch is supervised, fine-tuned for change-referred segmentation. The enhanced $T_v$ and foreground prior $Q_d$ (denoted as $T_p$) are then mapped via two-layer MLPs $P_v$ to serve as visual inputs for the LLM.

Finally, the LLM processes all inputs to generate text responses and segmentation tokens. As shown in Fig.~\ref{fig:framework}, it produces a temporal transition sequence (e.g., \texttt{[SEG]} to \texttt{[SEG]}, defaulting to $T_1 \rightarrow T_2$ in bi-temporal cases) to articulate evolutionary states. Upon generating a \texttt{[SEG]} token, an MLP $P_t$ maps its hidden states to the visual space~\cite{lai2024lisa,rasheed2024glamm}, which the mask decoder leverages alongside corresponding temporal features to predict a high-precision, pixel-level change mask.

\subsection{Local Causal Attention}
Applying standard autoregressive causal attention to multi-temporal inputs introduces an unwarranted positional bias, disrupting the ``fair comparison'' between temporal features (swapping input order degrades performance; see Appendix~B). To mitigate this, we propose Local Causal Attention (LCA). Building upon the standard mask ($M^{A}_{i,j} = -\infty$ for $j > i$), LCA introduces a phase-aware constraint: $M^{A}_{i,j} = -\infty$ if $i$ and $j$ are visual tokens from different temporal phases. This blocks cross-temporal feature leakage and generalizes seamlessly to $K$ frames or interleaved text prompts, while subsequent text tokens retain global causal attention. Furthermore, because the Change-SEG module spatially concatenates features $T_1$ to $T_K$ along the width dimension, flattening them yields an interleaved 1D sequence. Applying LCA to this sequence naturally forms a distinctive checkerboard-like attention pattern (Fig.~\ref{fig:framework}), ensuring phase-exclusive visual attention while preserving perfect spatial alignment.

\subsection{Training Strategy and Objective Functions}
Delta-LLaVA is optimized via a two-stage training strategy that progressively transitions from foundational single-view visual perception to advanced multi-temporal contrastive reasoning.

\textbf{Stage 1: Visual Alignment Pretraining.} This stage equips visual perception with fundamental difference-extraction capabilities. Keeping the LLM parameters frozen, we strictly fine-tune the visual components—specifically, the change enhanced module, the change decoder, and the projector. The objective function here comprises the text generation loss ($\mathcal{L}_{\text{text}}$) and the projection alignment loss ($\mathcal{L}_{\text{reg}}$):
\begin{equation}
    \mathcal{L}_{\text{pretrain}} = \mathcal{L}_{\text{text}} + \mathcal{L}_{\text{reg}}, \quad \mathcal{L}_{\text{reg}} = \| T_{ov} - P_t(P_v(T_{p})) \|_2^2.
\end{equation}

\textbf{Stage 2: Multi-Temporal Instruction Tuning.} Building upon the optimized visual modules, we additionally employ Low-Rank Adaptation (LoRA) to fine-tune the LLM, enabling it to deeply comprehend difference features and perform multimodal interactive reasoning. The total loss function synergistically combines the text generation loss ($\mathcal{L}_{\text{text}}$), the segmentation loss ($\mathcal{L}_{\text{mask}}$) formulated using Dice and Focal losses, and the label classification loss ($\mathcal{L}_{\text{cls}}$):
\begin{equation}
    \mathcal{L}_{\text{instruction}} = \mathcal{L}_{\text{text}} + \mathcal{L}_{\text{mask}} + \mathcal{L}_{\text{cls}}, \quad     \mathcal{L}_{\text{mask}} = \alpha \mathcal{L}_{\text{Focal}} + \beta \mathcal{L}_{\text{Dice}}.
\end{equation}
\begin{table*}[htbp]
    \centering
    \renewcommand{\arraystretch}{0.9}
\caption{Semantic change detection (SCD) results on Delta-SECOND. Best scores for specialized and reasoning models are shaded separately. Abbreviations: L. Veg. (Low Vegetation), Grnd. (Ground), Build. (Building), and Play. (Playground). * denotes fine-tuned baselines.}    
    \vspace{-0.1cm}
\definecolor{TopBest}{HTML}{BFBFBF}  
    \definecolor{TopSec}{HTML}{E6E6E6}   
    \definecolor{BotBest}{HTML}{9BC2E6}  
    \definecolor{BotSec}{HTML}{DDEBF7}   
    \newcommand{\topbest}[1]{\cellcolor{TopBest}#1}
    \newcommand{\topsec}[1]{\cellcolor{TopSec}#1}
    \newcommand{\botbest}[1]{\cellcolor{BotBest}#1}
    \newcommand{\botsec}[1]{\cellcolor{BotSec}#1}
    \label{tab:scd_results}
    \resizebox{0.74\textwidth}{!}{
    \begin{tabular}{l cccc cccccc}
        \toprule
        \multirow{2}{*}{Methods} & \multicolumn{4}{c}{SCD Overall Metrics} & \multicolumn{6}{c}{Non-Background Class IoU} \\
        \cmidrule(lr){2-5} \cmidrule(lr){6-11}
        & mIoU & OA & SeK & $F_{scd}$ & L. Veg. & Grnd. & Tree & Water & Build. & Play. \\
        \midrule
        \multicolumn{11}{l}{\textit{Specialized SCD Models}} \\
        \midrule
        SAM-DINO-SegEarth-OV\cite{li2026dynamicearth}    & 60.85 &   91.74   &   13.53   & 33.88 &  21.14&  24.31&  13.37&  13.31&  35.26&  17.69\\
        SAM-DINOv2-SegEarth-OV\cite{li2026dynamicearth}  & 63.59 &   \topbest{92.51}   &   15.24   & 37.51 &  21.36&  27.95&  15.47&  15.15&  38.79&  22.85\\
        SAM2-DINOv2-SegEarth-OV\cite{li2026dynamicearth} & 61.86 &   \topsec{92.11}   &   14.26   & 35.33 &  20.05&  19.80&  16.45&  14.30&  36.78&  23.81\\
        ChangeMamba\cite{chen2024changemamba}                &   \topsec{72.85}&   88.16&   \topsec{21.55}&   \topsec{54.81}&  \topsec{32.49}  &  \topsec{41.17}  &  \topsec{27.62}  &  \topsec{37.81}  &  \topsec{54.83}  &  \topsec{36.42}  \\
        FoBa\cite{zhang2025foba}                &   \topbest{73.44}&   89.73&   \topbest{23.78}&   \topbest{56.96}&  \topbest{34.03}  &  \topbest{42.48}  &  \topbest{29.86}  &  \topbest{38.73}  &  \topbest{57.32}  &  \topbest{37.60}  \\
        \midrule
        \multicolumn{11}{l}{\textit{Reasoning Segmentation Models}} \\
        \midrule
        EVOL-SAM3\cite{ye2025evolving}            &   45.33&   86.68&   0.11&   2.15&  0.01&  0.57&  0.14&  \botsec{0.82}&  1.76&  \botsec{0.59}\\
        PSALM\cite{zhang2024psalm}*                & 43.22&   89.37&   0.01&   0.01&  0.02&  0.02&  0.00&  0.00&  0.01&  0.00  \\
        OMG-LLaVA\cite{zhang2024omg}*           & \botsec{47.11}& 90.29& \botsec{0.89}&  \botsec{7.00}&  \botsec{1.16}&  \botsec{1.89}&  \botsec{0.41}&  0.57&  \botsec{5.68}&  0.50\\
        LISA\cite{lai2024lisa}*                 & 46.06& \botsec{91.04}& 0.03&   0.04&  0.02&  0.01&  0.01&  0.00&  0.04&  0.00\\
        LISAT\cite{quenum2025lisat}*                & 46.04& \botsec{91.04}&  0.02&  0.02&  0.01&  0.01&  0.00&  0.00&  0.32&  0.03\\
        \midrule
        Delta-LLaVA (Ours)& \botbest{69.72} & \botbest{92.42} & \botbest{19.18} & \botbest{53.51} & \botbest{32.77}& \botbest{39.80}& \botbest{20.09}& \botbest{24.35}& \botbest{54.68}& \botbest{34.57}\\
        \bottomrule
    \end{tabular}
    }
\end{table*}

\begin{table*}[htbp]
    \centering
    \caption{Change QA performance of VLMs on Delta-SECOND. Metrics include Accuracy for choice questions (CIC), and METEOR (M) / CIDEr (C) for open-ended questions (CQS, CTI, CSA). * denotes fine-tuned baselines.}
    \renewcommand{\arraystretch}{0.9}
    \vspace{-0.1cm}
    \label{tab:reasoning_results}
    \resizebox{0.93\textwidth}{!}{
    \begin{tabular}{l|cccc|cccccccccccc}
        \toprule
        \multirow{3}{*}{Methods} & \multicolumn{4}{c}{Choice QA (Accuracy)} & \multicolumn{12}{c}{Open-Ended QA (METEOR \& CIDEr)} \\
        \cmidrule(lr){2-5} \cmidrule(lr){6-17}
        & \multicolumn{2}{c}{H-CIC} & \multicolumn{2}{c}{P-CIC} & \multicolumn{2}{c}{H-CQS} & \multicolumn{2}{c}{H-CTI} & \multicolumn{2}{c}{H-CSA} & \multicolumn{2}{c}{P-CQS} & \multicolumn{2}{c}{P-CTI} & \multicolumn{2}{c}{P-CSA} \\
        \cmidrule(lr){2-3} \cmidrule(lr){4-5} \cmidrule(lr){6-7} \cmidrule(lr){8-9} \cmidrule(lr){10-11} \cmidrule(lr){12-13} \cmidrule(lr){14-15} \cmidrule(lr){16-17}
        & Single & Multi & Single & Multi & M & C & M & C & M & C & M & C & M & C & M & C \\
        \midrule
        \multicolumn{17}{l}{\textit{Commercial Generalist MLLMs}} \\
        \midrule
        GPT-4.1\cite{openai2025gpt41api}              &   60.20&   31.38&   \underline{57.86}&   29.20&   16.40&   28.02&   15.77&   30.26&   14.21&   28.77&   17.22&   32.30&   16.80&   27.61&   15.13&   25.38\\
        GPT-4o\cite{hurst2024gpt}               &   57.71&   30.75&   48.72&   28.42&   15.79&   26.98&   18.25&   28.03&   14.89&   26.08&   15.83&   31.15&   18.57&   29.11&   15.14&   24.23\\
        Gemini 2.5 Flash\cite{comanici2025gemini}     &   \underline{61.69}&   31.00&   53.48&   29.07&   16.18&   27.21&   13.89&   29.64&   14.60&   27.51&   17.15&   30.45&   15.35&   30.74&   15.67&   26.19\\
        \midrule
        \multicolumn{17}{l}{\textit{Open-source Generalist MLLMs}} \\
        \midrule
        Qwen2.5-VL 7B\cite{Bai2025Qwen25VLTR}        &   59.70&   22.38&   42.46&   21.58&   13.26&   22.40&   16.58&   20.03&   11.81&   22.85&   13.52&   20.36&   16.61&   20.04&   11.87&   20.07\\
        Qwen3-VL 8B\cite{Bai2025Qwen25VLTR}          &   52.24&   24.25&   38.56&   23.26&   15.47&   24.11&   15.70&   24.38&   14.76&   23.10&   15.60& 22.92  &   16.57&   21.17&   15.24&   21.78\\
        InternVL2.5 8B\cite{chen2024expanding}       &   56.22&   23.13&   32.67&   20.16&   10.44&   22.06&   10.57&   23.19&   13.26&   21.74&   12.21&   24.21&   15.49&   22.01&   13.28&   20.89\\
        InternVL3 8B\cite{zhu2025internvl3}         &   50.25&   25.50&   38.63&   25.32&   11.21&   24.35&   12.83&   24.82&   14.69&   22.51&   15.42&   24.65&   14.50&   23.19&   14.75&   21.97\\
        \midrule
        \multicolumn{17}{l}{\textit{Reasoning Segmentation Models}} \\
        \midrule
        EVOL-SAM3\cite{ye2025evolving}            &   47.76&   24.63&   39.63&   18.22&  14.51&   20.63&   13.44&   20.42&   15.59&   20.22&   14.56&   20.59&   14.88&   20.18&   14.55&   20.17\\
        PSALM\cite{zhang2024psalm}*            &   48.26&   29.25&   51.17&   30.10&  24.32&  28.45 &   21.43&   32.59&   22.73&   38.63&   26.18&   42.85&   23.60&   31.52&   21.36&   29.41\\
        OMG-LLaVA\cite{zhang2024omg}*           &   50.25&  31.25&   55.24   &   37.73  &   26.80&   37.06&  22.86&   34.73&   23.87&   41.26&   28.45   &   44.73   &   \underline{26.55}   &   32.98   &   23.89   &   30.97   \\
        LISA\cite{lai2024lisa}*           &   50.75&   37.75&   53.69&   37.73&   \underline{28.64}&   32.13&   23.77&   35.92&   \underline{26.58}&   \underline{41.88}&   \underline{29.96}&   43.43&   25.29&   \underline{33.84}&   \underline{25.53}&   \underline{33.58}\\
        LISAT\cite{quenum2025lisat}*                &   57.71&   \underline{39.88}&   56.63&   \underline{40.18}&   27.71&   \underline{37.07}&   \underline{23.88}&   \underline{40.21}&   26.35&   40.71&   29.59&   \underline{45.99}&   25.27&   33.73&   25.15&   31.67\\
        \midrule
        Delta-LLaVA (Ours)      &\textbf{64.18}&\textbf{43.38}&\textbf{60.70}&\textbf{41.34}&\textbf{32.74}&\textbf{41.45}&\textbf{27.77}&\textbf{45.00}&\textbf{30.31}&\textbf{42.80}&\textbf{31.93}&\textbf{54.01}&\textbf{29.49}&\textbf{40.25}&\textbf{29.91}&\textbf{37.16}\\
        \bottomrule
    \end{tabular}
    }
    \vspace{-0.1cm}
\end{table*}
\section{Experiments}
\subsection{Experimental Settings}
\paragraph{Baselines.} We compare Delta-LLaVA against 15 models across three categories at their default resolutions: (1) \textit{Reasoning Segmentation Models}: LISA~\cite{lai2024lisa}, OMG-LLaVA~\cite{zhang2024omg}, LISAT~\cite{quenum2025lisat}, PSALM~\cite{zhang2024psalm}, and EVOL-SAM3~\cite{ye2025evolving}; (2) \textit{Specialized SCD Models}: SAM-DINO-SegEarth-OV series~\cite{li2026dynamicearth}, ChangeMamba~\cite{chen2024changemamba}, and FoBa~\cite{zhang2025foba}; (3) \textit{Generalist MLLMs}: Qwen2.5/3-VL~\cite{Bai2025Qwen25VLTR,yang2025qwen3}, InternVL2.5/3~\cite{chen2024expanding,zhu2025internvl3}, GPT-4o/4.1~\cite{hurst2024gpt,openai2025gpt41api}, and Gemini 2.5 Flash~\cite{comanici2025gemini}. 

\paragraph{Evaluation Metrics.} For pixel-level change detection, we report $mIoU$, Overall Accuracy ($OA$), $SeK$ (Separable Kappa), $F_{scd}$ (SCD F-score), and category-specific $IoU$. For change understanding QA, we measure Accuracy (single- and multiple-choice) alongside METEOR and CIDEr (open-ended).

\paragraph{Implementation Details.} Delta-LLaVA employs a ConvNeXt-L CLIP encoder~\cite{liu2022convnet,radford2021learning} and InternLM2-7B~\cite{cai2024internlm2} (max length 8192). Models are trained on 8 NVIDIA A800 GPUs using FP16 mixed precision and DeepSpeed ZeRO-2. Training uses a two-stage strategy: first optimizing the projector and segmentation module, followed by LLM LoRA fine-tuning ($r=256$). For fair comparison, reasoning segmentation baselines are supervised fine-tuned on Delta-QA. Zero-shot generalization is evaluated on the BCA-QA task of DVL-Bench~\cite{xuan2025dynamicvl}.

Detailed metrics, hyperparameters, and configurations are introduced in Appendix~B.

\subsection{Main Results}
This section evaluates models on the Delta-SECOND and Delta-WUSU subsets, and tests generalization on DVL-Bench~\cite{xuan2025dynamicvl}. Detailed Delta-Landsat results are provided in Appendix~B.

\subsubsection{Semantic Change Detection Results}
Table \ref{tab:scd_results} details SCD performance on Delta-SECOND, revealing two key insights. First, \textbf{reasoning segmentation baselines exhibit systemic temporal failure.} Despite supervised fine-tuning, standard reasoning segmentation models (e.g., LISA, LISAT) notably fail, with $F_{scd}$ remaining near zero. This proves that without intrinsic multi-temporal comparison mechanisms, static models cannot simply ``learn'' mask-level change detection. Conversely, Delta-LLaVA successfully bridges temporal perception and spatial grounding, achieving a robust $69.72\%$ mIoU. Second, \textbf{Delta-LLaVA successfully unifies fine-grained change perception with rich semantic reasoning.} Compared to pure-vision SCD experts (e.g., FoBa, ChangeMamba), Delta-LLaVA has closed the gap in pixel-level masking. This confirms its capability for the fine-grained perception of visual feature changes. Furthermore, while specialized models are rigidly confined to generating closed-set masks, Delta-LLaVA simultaneously delivers rich language explanation capabilities, effectively integrating precise mask-level localization with open-ended visual-language understanding.

\subsubsection{Change Understanding Question Answering Results}
Table \ref{tab:reasoning_results} demonstrates that \textbf{Delta-LLaVA achieves state-of-the-art performance} across all QA tasks on the Delta-SECOND subset. The results reveal two critical insights. \textbf{First, generalist MLLMs exhibit multi-temporal blindness.} Leading models (e.g., GPT-4.1, Gemini 2.5 Flash) manage moderate accuracy on simple single-choice questions but notably fail on complex open-ended tasks (CIDEr typically $<33.0$). This exposes their lack of intrinsic comparative mechanisms to comprehend evolutionary dynamics. \textbf{Second, explicit change modeling drastically outperforms naive fine-tuning.} Despite supervised fine-tuning on our dataset, reasoning segmentation baselines (LISA, LISAT) are decisively beaten. For example, Delta-LLaVA surpasses the second-best LISAT by an impressive 8.02 CIDEr points on P-CQS and $>14\%$ in H-CIC multi-choice accuracy. This substantial gap validates that merely adapting static architectures is suboptimal, whereas Delta-LLaVA's dedicated temporal perception mechanisms successfully bridge the semantic gap between visual difference extraction and logical language reasoning.

\begin{figure*}[t]
  \centering
  \begin{minipage}{0.3\textwidth}
    \centering
    \includegraphics[width=\linewidth]{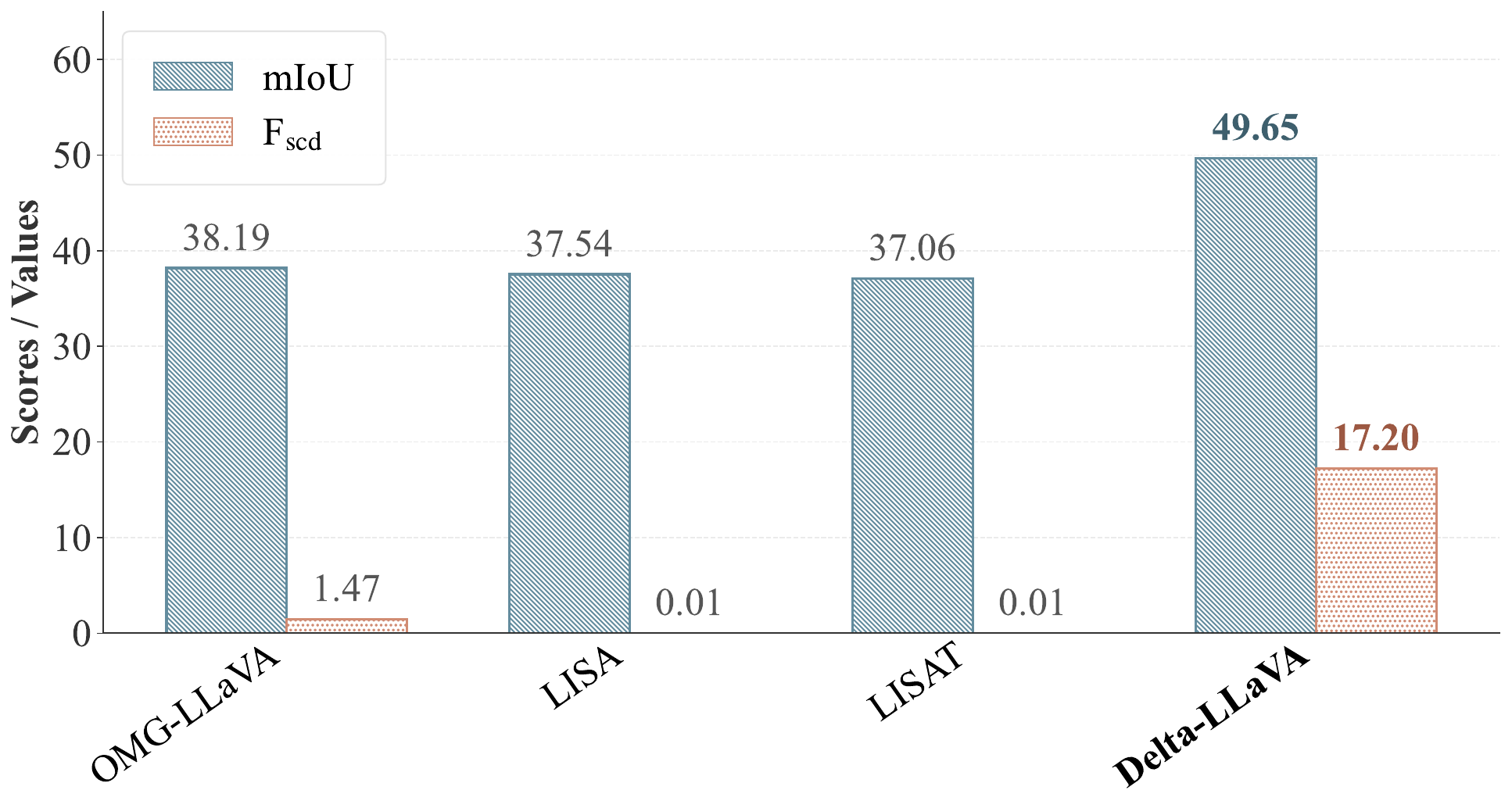}
    \caption{Performance of change segmentation on the Delta-WUSU.}
    \label{fig:wusu_scd}
  \end{minipage}\hfill
  \begin{minipage}{0.3\textwidth}
    \centering
    \includegraphics[width=\linewidth]{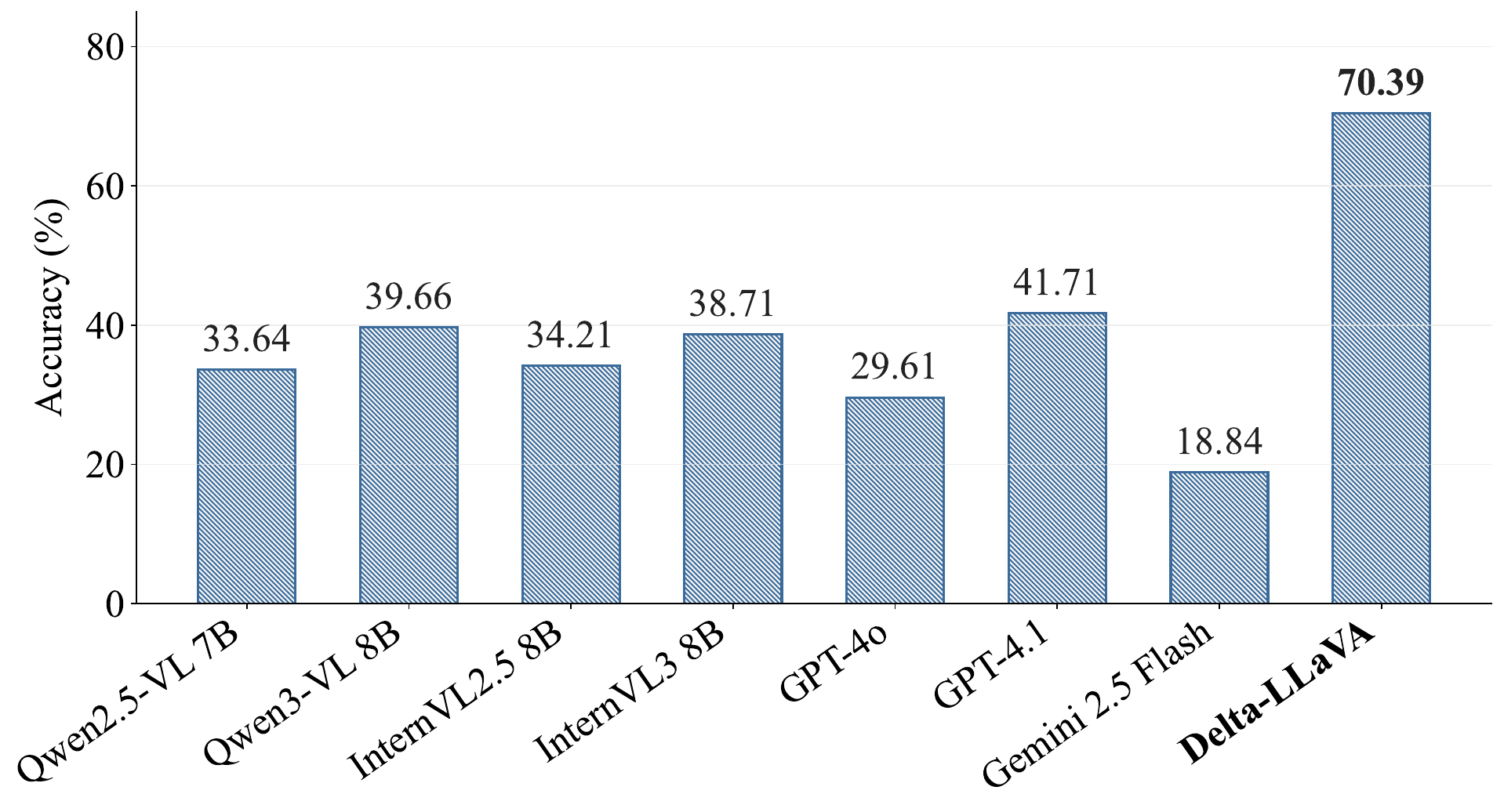}
    \caption{Performance of the QA task on the Delta-WUSU.}
    \label{fig:wusu_cq}
  \end{minipage}\hfill
  \begin{minipage}{0.3\textwidth}
    \centering
    \includegraphics[width=\linewidth]{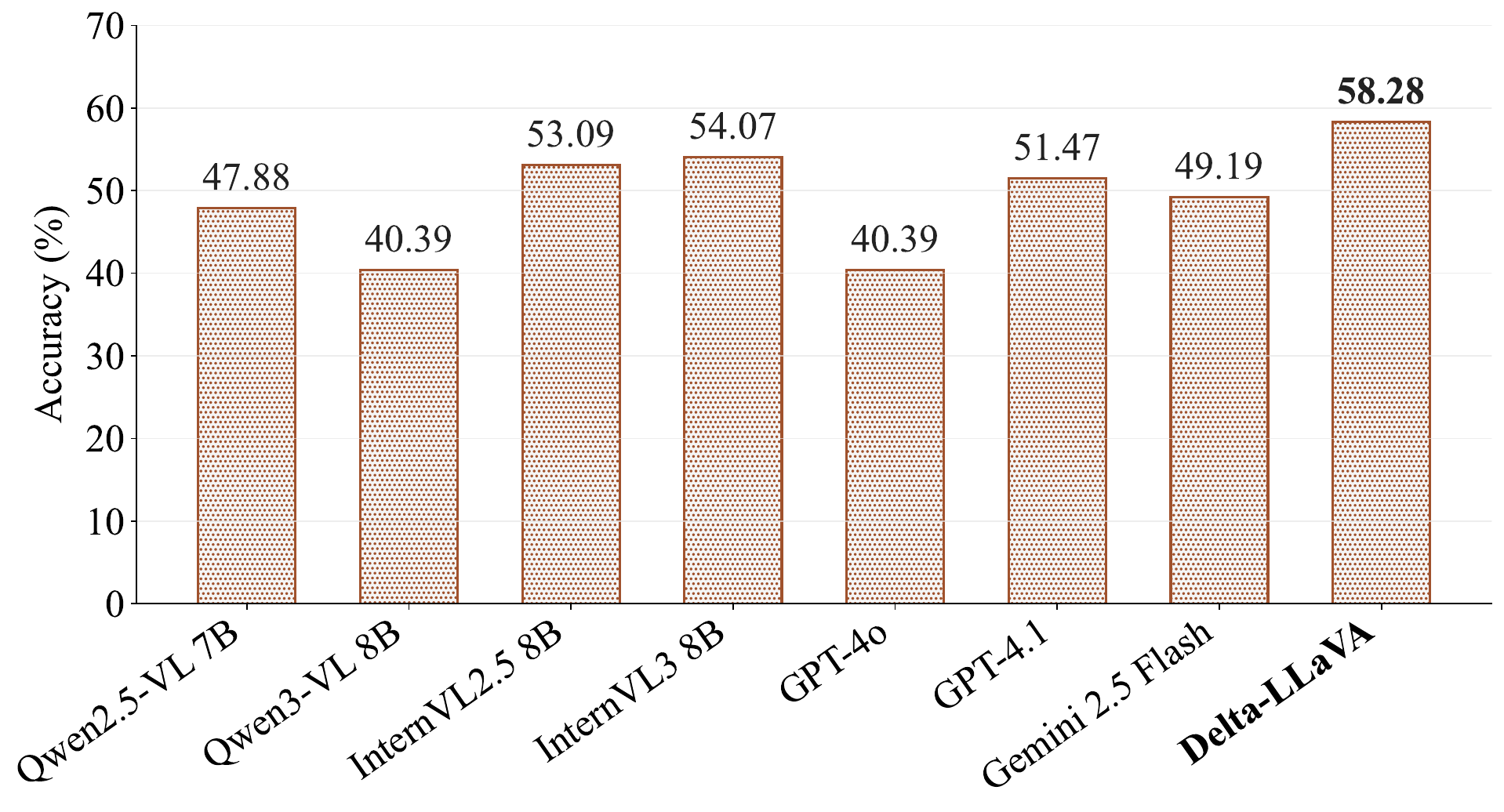 }
    \caption{Performance of the QA task on the BCA-QA subset of DVL-Bench.}
    \label{fig:dvl_cq}
  \end{minipage}
  \vspace{-0.4cm}
\end{figure*}

\begin{figure*}[htpb]
  \centering
  \includegraphics[width=\textwidth]{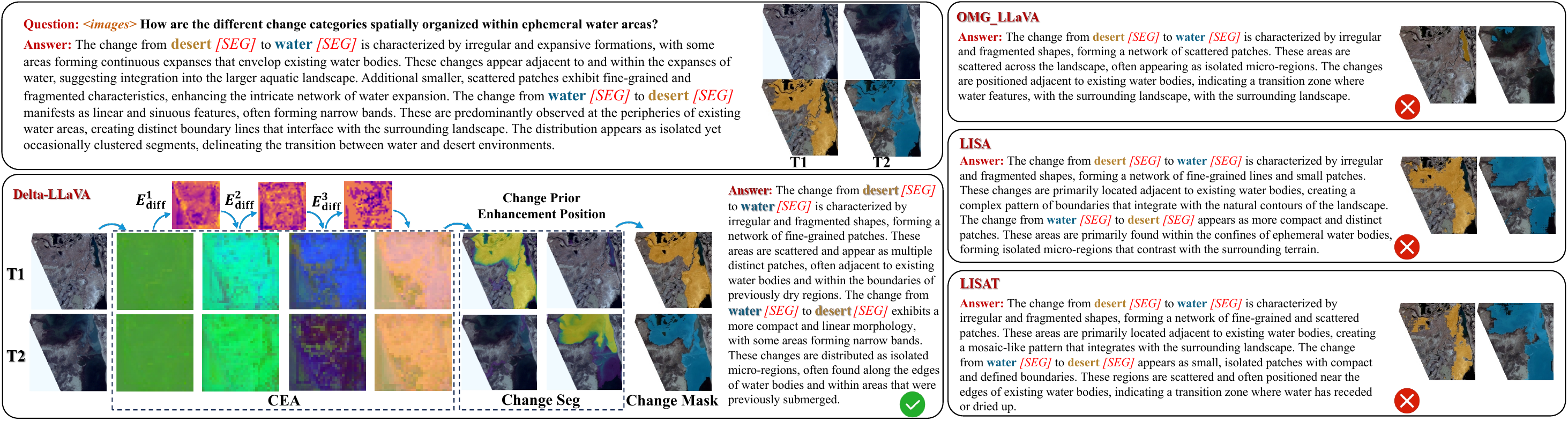}
  \caption{Qualitative comparison of Delta-LLaVA and other models for bi-temporal change understanding.}
  \label{fig:case1}
    \vspace{-0.2cm}
\end{figure*}

\subsubsection{Tri-temporal Analysis on Delta-WUSU}
We evaluate complex multi-step temporal deduction on the tri-temporal Delta-WUSU dataset. Extending the sequence to three phases ($T_1 \rightarrow T_2 \rightarrow T_3$) poses a significant challenge for existing architectures. In Semantic Change Detection (Fig. \ref{fig:wusu_scd}), reasoning segmentation baselines like LISA~\cite{lai2024lisa} and LISAT~\cite{quenum2025lisat} notably fail ($F_{scd} \sim 0.01\%$), whereas Delta-LLaVA maintains robust feature alignment, achieving 49.65\% mIoU and 17.20\% $F_{scd}$. This structural superiority extends to the QA task (Fig. \ref{fig:wusu_cq}). Lacking native multi-temporal mechanisms, generalist MLLMs (e.g., GPT-4.1~\cite{openai2025gpt41api}, InternVL3 8B~\cite{zhu2025internvl3}) fail to break a 42\% accuracy ceiling. In contrast, Delta-LLaVA reaches 70.39\%, confirming its unique ability to preserve change signals and execute complex reasoning across extended temporal contexts.

\subsubsection{Generalization Performance}
We evaluate out-of-domain generalization on the external DVL-Bench~\cite{xuan2025dynamicvl} (Fig.~\ref{fig:dvl_cq}). Delta-LLaVA achieves a leading $58.28\%$ accuracy on the BCA-QA task. It decisively outperforms universal commercial models, including GPT-4.1~\cite{openai2025gpt41api} ($51.47\%$) and Gemini 2.5 Flash~\cite{comanici2025gemini} ($49.19\%$), as well as the prevenient open-source competitor, InternVL3~\cite{zhu2025internvl3} ($54.07\%$). This confirms that our dedicated temporal alignment mechanisms provide more resilient and generalized change deduction capabilities than standard foundation models.

\begin{table}[htbp]
    \centering
        \vspace{-0.1cm}
    \caption{Ablation study of Delta-LLaVA model.}
    \label{tab:ablation}
    \large
    \resizebox{\columnwidth}{!}{
    \begin{tabular}{l cccc cc cc}
        \toprule
        \multirow{2}{*}{Model Variant} & \multicolumn{4}{c}{SCD Overall Metrics} & \multicolumn{2}{c}{Choice QA (Avg. Acc.)} & \multicolumn{2}{c}{Open-Ended QA (Avg.)} \\
        \cmidrule(lr){2-5} \cmidrule(lr){6-7} \cmidrule(lr){8-9}
        & mIoU & OA & SeK & $F_{scd}$ & Single & Multi & METEOR & CIDEr \\
        \midrule
        \textbf{Full Model} & \textbf{69.72} & \textbf{92.42} & \textbf{19.18} & \textbf{53.51} & \textbf{61.03} & \textbf{42.38} & \textbf{30.13} & \textbf{55.59}\\
        w/o CPE & 57.23 \scriptsize{\textcolor{red}{(-12.49)}} & 90.09 \scriptsize{\textcolor{red}{(-2.33)}} & 8.76 \scriptsize{\textcolor{red}{(-10.42)}} & 40.22 \scriptsize{\textcolor{red}{(-13.29)}} & 50.85 \scriptsize{\textcolor{red}{(-10.18)}} & 31.89 \scriptsize{\textcolor{red}{(-10.49)}} & 21.97 \scriptsize{\textcolor{red}{(-8.16)}} & 41.01 \scriptsize{\textcolor{red}{(-14.58)}} \\
        w/o CEA & 61.13 \scriptsize{\textcolor{red}{(8.59)}} & 90.71 \scriptsize{\textcolor{red}{(-1.71)}} & 11.19 \scriptsize{\textcolor{red}{(-7.99)}} & 44.38 \scriptsize{\textcolor{red}{(-9.13)}} & 55.29 \scriptsize{\textcolor{red}{(-5.74)}} & 35.45 \scriptsize{\textcolor{red}{(-6.93)}} & 26.31 \scriptsize{\textcolor{red}{(-3.82)}} & 44.12 \scriptsize{\textcolor{red}{(-11.47)}} \\
        w/o LCA & 64.32 \scriptsize{\textcolor{red}{(-5.40)}} & 91.23 \scriptsize{\textcolor{red}{(-1.19)}} & 15.50 \scriptsize{\textcolor{red}{(-3.68)}} & 48.91 \scriptsize{\textcolor{red}{(-4.60)}} & 56.83 \scriptsize{\textcolor{red}{(-4.20)}} & 36.04 \scriptsize{\textcolor{red}{(-6.34)}} & 27.26 \scriptsize{\textcolor{red}{(-2.87)}} & 46.81 \scriptsize{\textcolor{red}{(-8.78)}} \\
        \bottomrule
    \end{tabular}
    }
        \vspace{-0.3cm}
\end{table}
\subsection{Ablation Study}
Table \ref{tab:ablation} validates the essential contribution of each Delta-LLaVA component. The Change Prior Embedding (CPE) provides foundational difference representations; disabling it results in the most severe degradation ($-12.49\%$ mIoU, $-14.58$ CIDEr), demonstrating that it is a prerequisite for change understanding. Building on this, the Change-Enhanced Attention (CEA) critically amplifies these initial signals, as evidenced by drops of $8.59\%$ in mIoU and $11.47$ in CIDEr upon its removal. Finally, removing the Local Causal Attention (LCA) reduces mIoU by $5.40\%$ and CIDEr by $8.78$. This confirms the necessity of LCA for preserving causal relationships and preventing the erroneous confounding of cross-temporal visual features. Together, these modules cohesively bridge multi-temporal visual perception and language comprehension.

\subsection{Qualitative Results}
Fig.~\ref{fig:case1} presents a case study on bi-temporal spatial change analysis. As illustrated, Delta-LLaVA leverages the three-layer CEA to differentially amplify changed regions, with the attention heatmaps clearly highlighting the locations of these differences. Concurrently, the change queries within Change-SEG accurately capture the areas requiring enhanced difference priors. Consequently, Delta-LLaVA effectively accomplishes both change detection and understanding. In contrast, other baseline models struggle with either low segmentation accuracy or inaccurate textual descriptions, failing to properly comprehend the changes. Additional visualization results are provided in Appendix~C.

\begin{table}[htbp]
\centering
\renewcommand{\arraystretch}{0.85}
\caption{Model parameters and average computational time on 100 samples from Delta-SECOND.}
    \vspace{-0.1cm}
\label{tab:merged_efficiency}
\large
\resizebox{0.48\textwidth}{!}{
\begin{tabular}{lcccccc}
\toprule
\multirow{2}{*}{Component} & \multicolumn{4}{c}{Parameters} & \multicolumn{2}{c}{Computational Time} \\
\cmidrule(lr){2-5} \cmidrule(lr){6-7}
 & Total (M) & \% of Total & Trainable (M) & \% of Train. & Time (ms) & Ratio (\%) \\
\midrule
LLM & 5505.52 & 92.10 & 2015.59 & 88.82 & 13539.22 & 96.83 \\
Visual Encoder & 205.85 & 3.40 & 0.00 & 0.00 & 193.55 & 1.38 \\
Frozen Change-SEG & 14.93 & 0.20 & 0.00 & 0.00 & 142.01 & 1.02 \\
Trainable Change-SEG & 14.80 & 0.20 & 14.80 & 0.65 & 78.50 & 0.56 \\
CEA & 147.26 & 2.50 & 147.26 & 6.49 & 23.62 & 0.17 \\
Visual Projector & 89.16 & 1.50 & 89.16 & 3.93 & 5.15 & 0.04 \\
Text Projector & 2.36 & 0.04 & 2.36 & 0.10 & $<$ 0.10 & $\sim$ 0.00 \\
\bottomrule
\end{tabular}
}
    \vspace{-0.2cm}
\end{table}
\subsection{Computational Complexity Analysis}
To evaluate Delta-LLaVA's potential for lightweight spaceborne computing, Table~\ref{tab:merged_efficiency} analyzes its parameter distribution and inference time. While the foundational LLM naturally dominates computational costs ($92.10\%$ of parameters and $96.83\%$ of one A800 inference time), our core innovations introduce negligible overhead. Specifically, the trainable Change-SEG and CEA modules combined consume under $1\%$ of the total inference time and constitute merely $7.14\%$ of trainable parameters. This confirms that our dedicated change perception mechanisms achieve advanced multi-temporal alignment without imposing computational bottlenecks.

\section{Conclusion}

This paper presents a unified multimodal paradigm for remote sensing change detection and understanding. We introduce the Delta-QA dataset to standardize multi-temporal evaluation across four cognitive dimensions, alongside Delta-LLaVA, a novel architecture resolving the temporal alignment bottlenecks of existing MLLMs. Driven by the VCP module and LCA, Delta-LLaVA accurately isolates temporal difference priors while eliminating cross-temporal feature leakage. Extensive experiments confirm that our model achieves state-of-the-art pixel-level segmentation and complex open-ended reasoning with negligible computational overhead. Future work will continue to overcome limitations by scaling to extended time series and refining MLLMs to match specialized SCD models.

\bibliographystyle{ACM-Reference-Format}
\bibliography{sample-base}


\end{document}